# Some Experiments with Real-time Decision Algorithms

Bruce D'Ambrosio
Computer Science
Oregon State University
Corvallis, Oregon 97331
dambrosi@research.cs.orst.edu

Scott Burgess
Computer Science
Oregon State University
Corvallis, Oregon 97331
burgess@research.cs.orst.edu

## Abstract

Real-time Decision algorithms are a class of incremental resource-bounded [Horvitz, 89] or anytime [Dean, 93] algorithms for evaluating influence diagrams. We present a test domain for real-time decision algorithms, and the results of experiments with several Real-time Decision Algorithms in this domain. The results demonstrate high performance for two algorithms, a decision-evaluation variant of Incremental Probabilisitic Inference [D'Ambrosio, 93] and a variant of an algorithm suggested by Goldszmidt, [Goldszmidt, 95], PK-reduced. We discuss the implications of these experimental results and explore the broader applicability of these algorithms.

## Introduction

### The problem

A variety of algorithms have been proposed as candidates for *anytime* [Dean, 93] or *resource-bounded* [Horvitz et al, 89] inference, including [D'Ambrosio, 93], [Horvitz et al, 89b], and a variety of simulation-based algorithms such as [Fung, 89]. The need for such algorithms arises because implementable agents have finite computing resources [Russell, 91]. The world in which an agent is embedded continues to evolve while the agent chooses an action. Thus, the utility of an action depends not only on the action selected, but also on the time at which the action is performed, which in turn depends on how long it takes the agent to choose the action. In these circumstances, a fast but approximate decision algorithm may outperform an "optimal" but slower one. In this paper we present experimental results characterizing several promising candidate real-time decision algorithms. We begin with a short review of the set of algorithms we chose to characterize: a search-based algorithm (Incremental Probabilistic Inference, [D'Ambrosio, 93]) and two variants of a decision algorithm suggested by Goldszmidt [Goldszmidt, 95]. Characterizing such algorithms is non-trivial. We describe the On-Line Maintenance Agent (OLMA) [D'Ambrosio, 92], [D'Ambrosio, 96], an idealized task domain that has the necessary properties to permit informative experimental estimation of the performance properties of the various algorithms. We then present experimental results obtained using each of the test algorithms (and two reference algorithms, exact inference and random choice) on a sample problem in the OLMA domain. We close with a discussion of the results.

Our primary findings are, first, that real-time algorithms do indeed make sense in this domain and, second, that the best algorithms exhibit a smooth tradeoff between time spent and quality of decision. Our experimental evidence supports our intuition that, as more time is available, it pays to "think" more deeply before acting. The algorithm with the best overall performance is one of the variants of the Goldszmidt algorithm, although we will place some caveats on this conclusion in the discussion section.

## The Candidate Algorithms

Our evaluation focused on two promising approximate decision algorithms we term D-IPI and K-reduced. D-IPI is an extension of the IPI search algorithm [D'Ambrosio, 93] to include decision and value nodes. K-reduced is a use of Goldszmidt's fast method of computing prior Kappa values [Goldszmidt, 95]. In this section we briefly describe each of these algorithms, as well as several reference algorithms we used to establish benchmark solution values.

### D-IPI

D-IPI is a simple extension of the IPI incremental inference algorithm [D'Ambrosio, UAI-93]. IPI is an incremental search-based variant of the SPI [Li & D'Ambrosio, 94] algorithm. It first forms a symbolic expression (marginalization over the joint pdf) corresponding to a query. It then constructs an evaluation tree for the query by applying simple algebraic transforms to convert the expression into efficiently evaluable form. Finally, it searches the tree top-down for large-value joint instantiations of the variables. IPI uses caching to identify repeated visits to a tree node, and dependency tracking to update all parents when a subtree is searched further. We have shown that, through these techniques, IPI retains the space and time complexity of efficient exact inference algorithms. The IPI algorithm as described in [D'Ambrosio, 93] searches over evaluation trees consisting of conformal product operations. It is a simple extension to enable IPI to search over more general expressions, including sum and difference operators. This yields an algorithm capable of performing incremental inference over our full local expression language [D'Ambrosio, UAI-91; D'Ambrosio, IJAR-95]. D-IPI is a further extension of the IPI algorithm to include a maximization operator. This yields an algorithm capable of searching over MSEU expressions. Construction of an incremental form of the maximization operator is an interesting programming exercise, but presents no interesting theoretical challenges.

We should note that the factoring algorithm used in IPI is quite different from that used in normal SPI. The goal of factoring for SPI is to minimize the size of the largest intermediate conformal product. This is one of the goals for factoring an expression for IPI, but a second, equally important goal, is to place highly skewed distributions early in the search process. Finally, the version of D-IPI used in these experiments is a relatively simple one that makes no attempt to optimize maximization operators: we simply form the expression for expected utility and then repeatedly maximize and marginalize, working back from the last decision in the diagram.

### K-reduced

Goldszmidt [Goldszmidt, UAI-95] has presented an algorithm for rapid computation of prior Kappa values in a belief net. In that paper, he suggested that this algorithm could be used to compute reduced domains for the variables in a network (i.e., for each variable select only those domain values with Kappa = 0), and that exact inference over these reduced domains might be an interesting form of approximate inference. We implemented a variant of this technique as follows: rather than actually compute Kappa values, we simply compute prior probabilities for every node in the network ignoring loops. That is, given a node ordering, we compute for each node in order:

$$P(n_i) = P(n_i | \pi_i) \Pi_{j \in \pi_i} P(n_{ij})$$

where $\pi_i$ is the set of immediate parents of node $i$. This step can be performed in time linear in the number of nodes in the network. We then build a list of all probabilities thus computed, in descending order (duplicates eliminated). This step takes $n \log(n)$ time. Finally, we select the highest probability value computed for each node, and then select the smallest value in this set (the *least-greatest-prior*). We then construct an anytime algorithm as follows:

 1. Find the least-greatest-prior in the sorted list of all priors. Call this the *current-minimum-prior*.

 2. For as long as you like, iterate:

  2.1 For each node in the network, reduce its domain to include only those values whose prior is greater than or equal to the current-minimum prior. Notice that, by construction, every node will have at least one such domain value.

  2.2 Apply your favorite decision algorithm to the resulting network. This yields the decision recommendation for this iteration.

2.3 Replace the current-minimum-prior with the next smaller entry in the sorted list of priors.

We call this algorithm Kappa-reduced Exact, or K-reduced, even though we don't explicitly compute Kappa values, since it is essentially identical to the procedure described by Goldszmidt in conversation.

### PK-reduced

As will be seen shortly, experimental results showed mediocre performance for K-reduced. This led us to try a variant in which we estimate *posterior* probabilities, rather than priors. We call this variant PK-reduced. This variant makes two changes to the K-reduced algorithm. First, we reduce the distributions at all evidence nodes and each of their immediate children by selecting only values consistent with the evidence. Then we perform the prior estimation described above. Finally, we perform a sweep *back* through the net, starting at evidence nodes[1]. That is, for each parent of each evidence node, we compute $P(p) \Pi_j P'(e_j | p)$.

This computation proceeds backward through the net in a manner analogous to $\Lambda$ message propagation. In fact, some thought should make it clear that this procedure is, in fact, a sloppy variant of Pearl's polytree propagation algorithm.[2]

Once posteriors are estimated, we sort them into descending order and use them to reduce variable domains as described in the K-reduced algorithm description above.

## Random

As a benchmark, we used a random choice of action. We surmised there might be circumstances in which "doing something, anything" might be better than spending any time computing, or that spending a short time computing might invariably lead to choosing exactly the wrong action. Random provides a lower bound on available performance.

## Exact

Finally, we used the SPI [Li and D'Ambrosio, 94] algorithm, extended for decision evaluation, to compute exact action recommendations. This algorithm provides another benchmark against which to evaluate the real-time decision algorithms: there is no point using a real-time algorithm in those areas of parameter space where it performs no better than exact inference.

---

[1] This is not necessary if all evidence is at root nodes. However, for the sensor-based applications we focus on, evidence is typically at leaf nodes.

[2] It would be better to use the full polytree algorithm, and there is no reason not to do so. However, due to the topology of the particular networks used in our experimental evaluation, that would not change the experimental results we present in this paper.

## The Task: On-Line Maintenance

We chose the task of diagnosis and repair for evaluating our candidate algorithms. Diagnosis is often formulated as a static, detached process, the goal of which is the assessment of the exact (or most probable) state of some external system. In contrast, we view diagnosis as a dynamic, practical activity by an agent engaged with a changing and uncertain world. Further, we extend the task to include the repair task to focus diagnostic activity. Our initial investigations have focused on the task of diagnosing a simple digital system *in situ*. Our formulation of embedded diagnosis has the following characteristics:

- The equipment[3] under diagnosis continues to operate while being diagnosed.
- Multiple faults can occur (and can continue to occur after an initial fault is detected).
- Faults can be intermittent.
- There is a known fixed cost per unit time while the equipment is malfunctioning (i.e., any component is in a faulted state).
- The agent senses equipment operation through a set of fixed sensors and one or more movable probes.
- Action alternatives include probing test points, replacing individual components, and simply waiting for the next sense report. Each action has a corresponding cost.
- The agent can only perform one action at a time.
- The overall task is to minimize total cost over some extended time period during which several failures can be expected to occur.

We term this task the *On-Line Maintenance* task, and an agent intended for performing such a task an *On-Line Maintenance Agent* (OLMA). An interesting aspect of this reformulation of the problem is that diagnosis is not a direct goal. A precise diagnosis is neither always obtainable nor necessary. Indeed, it is not obvious *a priori* what elements of a diagnosis are even relevant to the decision at hand.

Our first commitment is that the task is essentially a decision-theoretic one. That is, the essential task of the agent is to *act* in the face of limited information. In order to formulate this problem decision-theoretically, the agent must have knowledge of several parameters of the situation: It must know the cost of each type of replacement or probe act, the cost of system outage, and expected probabilities of component failures over the next decision cycle[4]. The latter two costs will vary with agent processing capacity, since a slower agent will take longer to make a decision. This will increase the chance of a component failure during a single decision cycle, and increase the cost of a system outage over a decision cycle. A naive attempt to formulate this task decision-theoretically encounters three problems.

First, a proper decision-theoretic consideration of this task would require looking ahead over all decisions over the entire operational life of the equipment in order to optimize the first decision. This is clearly computationally infeasible, at least on-line. Second, even if the first problem can be solved, time is passing while the agent is computing the first action, and it is not clear how the agent should trade quality of a decision for timeliness of the solution in choosing actions. Finally, the agent must act repeatedly, yet each action is in a new context: not only must a new set of input data be considered, but also a new set of beliefs about system state, based on prior information and computation.

The infinite look-ahead problem can be broken into two subproblems, one for replacement actions and another for probe actions. We circumvent the first subproblem, that of infinite look-ahead for replacement actions, as follows. For replacement actions we use an assumption of policy stability to derive long term utilities for these actions. This assumption is roughly as follows: If I choose not to replace a component now, then, all other things being equal (i.e., no new unexpected sense data), I will make the same choice next time. Under this assumption, the temporal consequences of a decision extend, not for a single sense/act cycle, but several decision cycles into the future. This effectively translates into a multiplier for the equipment downtime cost. The equipment/agent pair retains interesting behavior as long as the multiplier, $t$, obeys the following constraints:

$t \gg r/f$

$t \ll r/pf$

where:

1. $t$ is the outcome state duration (effectively, the multiplier for failure costs),
2. $r$ is the cost of component replacement,
3. $f$ is the cost of component failure for a unit clock time, and

---

[3] We will use *system* or *agent* to refer to our diagnostic system, and *equipment* to refer to the target physical system.

[4] Not strictly true: one could formulate the problem as a model-free reinforcement learning problem and address it with decision-theory grounded algorithms, such as various forms of asynchronous dynamic programming (e.g., Q-learning).

4. *p* is the probability of component failure during a unit time interval.

For further discussion of these constraints see [D'Ambrosio, 92].

We resolve the second subproblem, that of determining the expected value of probe actions, by using the standard decision-theoretic heuristic of one-step look-ahead. The result of these two techniques gives us an abstract decision-basis for the first decision as shown in Figure 1, where the link from the state at time one to the value node reflects the costs of operating in that state for one decision cycle, and the link from the state at time two to the value node reflects the cost of operating in that state for *t* time units. We follow standard influence diagram notation in this figure: circles represent abstract stater node, rectangles represent decision nodes, and the rounded corner object is the utility node. The dashed arrows between *O0* and *D0* and between *O1* and *D1* indicate information arcs. Finally, we label the third state node *Sn* rather than *S2* to indicate is represents the long term consequences of the second decision, as described above. Note this is an *abstraction* of the actual decision basis used. *S0*, for example, actually contains 6 nodes for the 4 gate circuit studied in this paper, and *O0* contains 4 nodes.

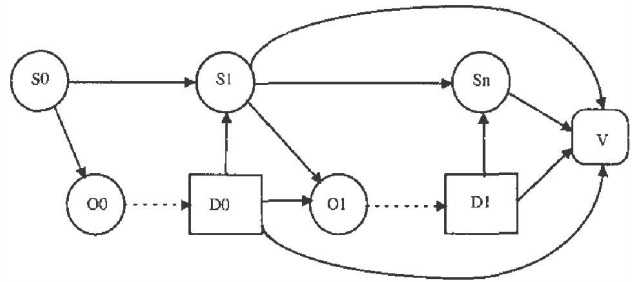

Figure 1: Abstract Influence Diagram for First Decision

Our second problem was that of trading quality-of-solution against time-to-solution. There are two issues here. First, if the equipment is faulted, the longer we delay taking a repair action, the higher the cost incurred. Second, since equipment operation is in parallel with agent operation, a fault may occur *while* the agent "wastes" time reasoning about a prior set of sense data. For this set of experiments we adopt a very simple agent architecture: once the agent begins reasoning it ignores all further input until it has chosen an action and executed it.

We resolve our final problem, that of making subsequent decisions, by simply extending the above decision basis forward in time by one decision stage each cycle. A sample decision basis for the second decision made by the maintenance agent is shown in Figure 2. This method would seem to have a problem: one would expect that decision time (and space) would increase at least linearly with time. In fact, both time and space requirements are constant. We simply replace the previous decision stage with the factored joint across posterior component state. Details of this vary somewhat depending on the decision algorithm used. For all algorithms except random and exact we used IPI to estimate the factors of the updated prior. For exact we used exact inference to compute the factors of the updated prior.

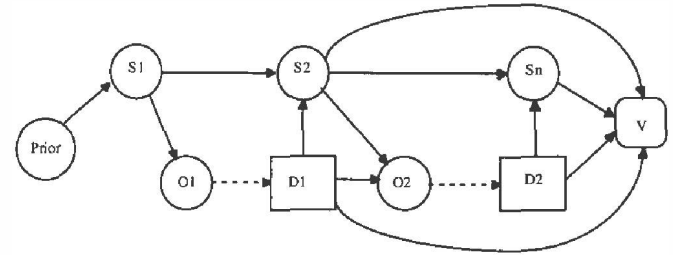

Figure 2: Influence Diagram for time 2

In summary, our agent executes the following cycle each time it is called upon to choose an action:

1. Extend the decision basis forward in time by one decision stage.
2. Acquire current sense data (including probe value if any).
3. Find the action with minimum expected cost.
4. Post the selected act as evidence in the belief net, prune (via posterior prior computation as discussed above) the now unneeded oldest stage from the net, and return selected action.

One final comment: the problem is surprisingly complex. The simple problem instance studied in this paper is well beyond the capability of current POMDP solution methods (the MDP state space for the simple 4 gate problem studied in this paper has 256 states, ignoring the stocahstic behaviour of the unknown mode!). Simple policies which only consider current observations can perform arbitrarily poorly

## Method

Our goal was to characterize the performance of the real-time algorithms with respect to variations in cpu speed. In particular, there are several hypotheses we wished to test:

1. The fundamental hypothesis on which both IPI and K-reduced are based is that it is possible to make effective decisions by considering only a very few instantiations of the decision model.

2. A further assumption of most real-time and anytime algorithm research is that it is in fact useful to vary the amount of computation performed as the time available (or equivalently, cpu speed) changes.

3. Finally, that there is a range of cpu speed over which the real-time algorithms outperform other alternatives, and become the decision method of choice.

Our experimental testbed has a "cpu-clock" parameter (*Quantum*) that controls the number of cpu seconds given the agent between each advance of the simulation clock. The greater the number of seconds given, the more time allowed for computation, and so the faster the effective speed of the cpu executing the decision algorithm. Each algorithm has a *step* parameter which controls the number of steps the algorithm should execute.

We designed test scenarios within the parameter space described earlier that would typically yield 7-10 component failures per scenario. In order to keep failure rates low enough this meant all runs were for at least 1000 simulation steps. We then adopted as our cost metric *cost per failure*, that is, total cost for a run divided by the number of failures which occurred. Each value shown in the graphs which follow is an average of at least two, and usually three, runs (each testbed run is made with a different random seed to generate a new pattern of component failures). Finally, the real-time algorithms (D-IPI, K-reduced, and PK-reduced) each have a parameter which must be set (number of terms to compute to D-IPI, and threshold probability to use for K-reduced and PK-reduced). We gathered data at each cpu-clock setting for a range of settings of these parameters, for each setting of the cpu-clock parameter. In all, several cpu-months of Sparc-2 time were consumed in gathering the data presented in the next section.

## Results

In this section we present numerical results of our experiments. We show detailed measurements for D-IPI, K-reduced, and PK-reduced, and overall results comparing all five algorithms. These results are discussed in the following section. We used the following parameters for all the runs in this section:

1. Failure probability (per gate): .003 (distributed uniformly among Stuck0, Stuck1, and Unknown, the three failure modes we modeled.)
2. Replacement cost: 3 per gate replaced
3. Probe cost: 1
4. Failure cost: 1 for each time step at least one gate is in a failure mode.

### D-IPI

Table 1 shows the numerical results obtained by averaging three runs for each point. The table shows Cost/Failure, the total cost of the run divided by the number of failures that occurred. Steps, for D-IPI, is the number of calls to the top of the search tree (number of terms or instantiations computed, sort of, see [D'Ambrosio, 93]). Missing entries in the table reflect missing data: we simply ran out of time to collect all the data needed.

| Steps | Quantum 1 | 2 | 4 | 8 | 16 | 32 | 64 | 128 |
|---|---|---|---|---|---|---|---|---|
| 1 | 40.7 | 27.8 | 25.9 | 20.56 | 26 | 29.5 | 21 | 20.6 |
| 2 | 37.3 | 34.4 | 28 | 27.3 | 28.7 | 21 | 26 | 22 |
| 4 | 40.3 | 27.7 | 23 | 22.9 | 21 | 19.8 | 21.5 | 19 |
| 8 |  | 41.2 | 35.9 | 25.4 | 23.3 | 30 | 19.2 | 35 |
| 16 | 65.6 | 72.8 | 60.5 | 23.4 | 23 | 38.3 | 28 | 22 |
| 32 | 59.2 | 63.6 | 48 | 24.8 | 21.9 | 18 | 23 | 29 |
| 64 |  | 40.1 | 50 | 19.75 | 22.6 | 19 | 19 | 20 |
| 128 |  | 107 | 63 | 32 | 23.5 | 15 | 18.6 | 18.8 |
| 256 |  | 122.7 | 86.5 | 58 | 45.5 | 24 | 24.8 | 21.4 |

Figure 3: D-IPI costs

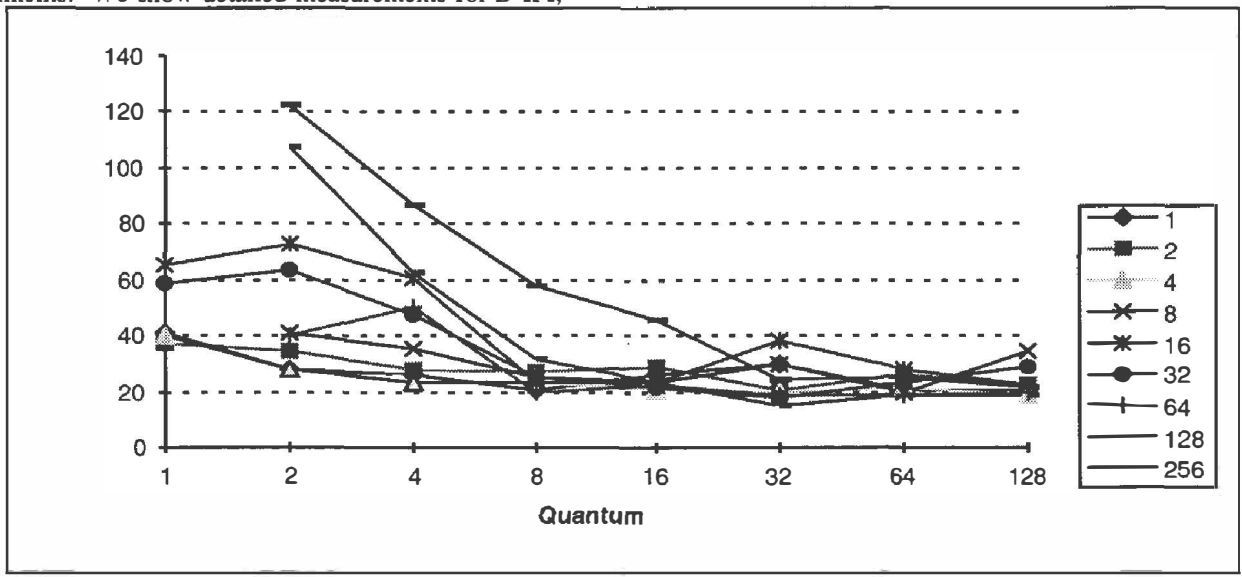

Figure 4: D-IPI Cost/Failure

### K-reduced

Next we present results for K-reduced. Preliminary data indicated that this algorithm was not competitive, so we did not collect a full data set for K-reduced.

|  | Steps |  |  |  |
|---|---|---|---|---|
| Quantum | 1 | 4 | 16 | 64 |
| 1 | 850 | 380 | 500 | 135 |
| 4 | 500 | 200 | 130 | 110 |
| 16 | 68 | 28 | 20 | 19 |
| 64 | 150 | 55 | 25 | 16.5 |

Figure 4: K-reduced cost/failure

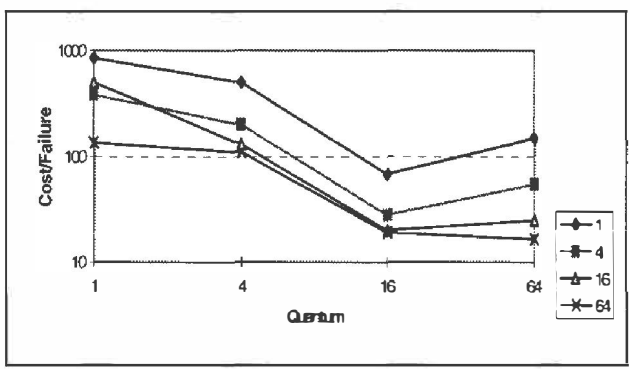

Figure 5: K-Reduced

### PK-reduced

We hypothesized that the reason for poor performance of K-reduced was that, since it was estimating *priors* to decide which values to include for each variable, it was ignoring the current evidence and making poor choices. A simple solution is to modify the algorithm to estimate *posteriors* instead, as described earlier. In this section we show the data collected for that modified algorithm. Figure 7 shows the cost/failure data for PK-reduced in tabluar form. the same data is showed in graphical form in figure 10.

|  | Quantum |  |  |  |  |  |  |
|---|---|---|---|---|---|---|---|
| # Steps | 2 | 4 | 8 | 16 | 32 | 64 | 128 |
| 1 | 41.4 | 33.1 | 23.9 | 21.8 | 28.4 | 24.3 | 21 |
| 2 | 21.3 | 34.5 | 20.6 | 17 | 18.6 | 17 | 19.2 |
| 4 | 25 | 26 | 20 | 38.9 | 19.95 | 18.9 | 18.6 |
| 8 | 29.7 | 20.3 | 22.1 | 21.1 | 17 | 18.34 | 18.1 |
| 16 | 36.9 | 23 | 18.5 | 16.95 | 17.7 | 20.6 | 15.9 |
| 32 | 53 | 31.8 | 22.9 | 25.4 | 17.9 | 19.1 | 15.8 |
| 64 | 76.3 | 41.9 | 29.9 | 31.6 | 17.3 | 15 | 18.6 |
| 128 | 175.1 | 39.7 | 34 | 36.6 | 17.3 | 18.8 | 15.5 |
| 256 | 90.7 | 43 | 29 | 40.2 | 14.8 | 18 | 14.1 |
| 512 | 113.4 | 50 | 38.8 | 37.7 | 17.6 | 17.3 | 18.3 |

Figure 6: PK-reduced Cost/Failure

### Overall

Finally, we show the overall results for all algorithms, including random and exact. For the three incremental algorithms we plot, for each quantum, the lowest cost/failure achievable by that algorithm at that quantum. Notice in the earlier charts that the optimum (minimum cost/failure) number of steps varies with quantum. As a result, these curves are generally "flatter" than any single curve in the previous graphs.

|  | Quantum |  |  |  |  |  |  |  |
|---|---|---|---|---|---|---|---|---|
| Algorithm | 1 | 2 | 4 | 8 | 16 | 32 | 64 | 128 |
| Random | 70.8 | 70.8 | 70.8 | 70.8 | 70.8 | 70.8 | 70.8 | 70.8 |
| Exact |  | 74 |  | 37 |  | 23 |  | 18 |
| Search | 37.3 | 27.7 | 23 | 19.75 | 21 | 18 | 18.6 | 18.8 |
| Kappa | 68 |  | 28 |  | 20 |  | 16.5 |  |
| PKappa |  | 21.3 | 20.3 | 18.5 | 16.95 | 14.8 | 15 | 14.1 |

Figure 8: Overall Comparison of Algorithms

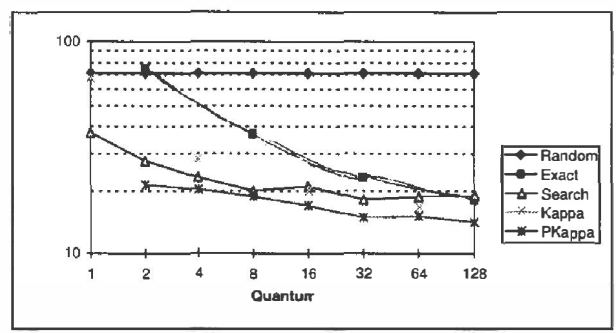

Figure 9: Overall Cost/Failure

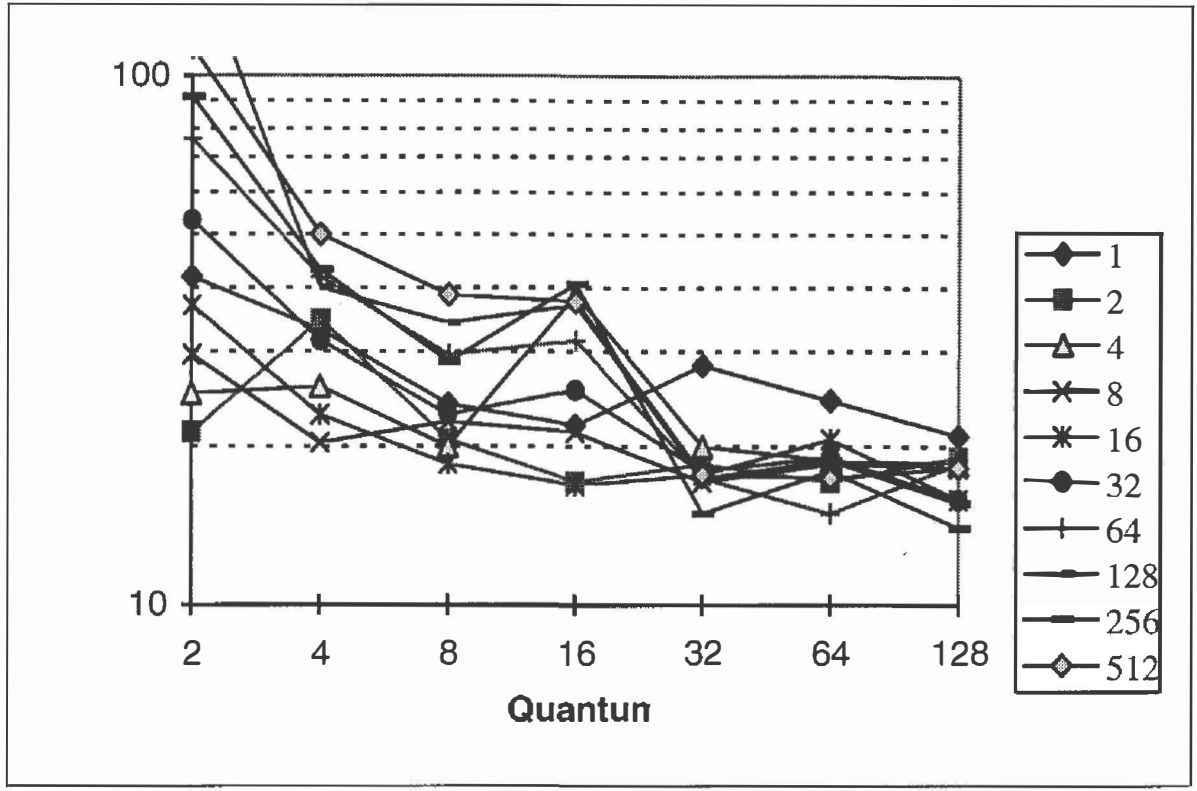

Figure 10: PK-Reduced

## Discussion

### OLMA results

All three real-time algorithms generally followed the expected trend:

    1. For a given number of steps of computation, the algorithm generally performs better when given more cpu time.

    2. The number of steps at which minimum cost/failure is obtained generally increases as more cpu time is available.

These results are consistent with, and support, the general theoretical framework for resource-bounded algorithms. Observation 2 is particularly interesting. While intuition predicts such results, it is reassuring to see that these algorithms do effectively trade cpu-time for solution quality in the step-size range of interest for real-time performance.

There are three interesting aspects of the results that we believe to be reproducible. First, we notice that for D-IPI and PK-reduced, very small amounts of computation, even a single increment, result in surprisingly good decision making. We are surprised that one can make reasonable decisions with so little computation, and are investigating the study problem further to understand why this is so.

One reason, we believe, is that the problem is relatively benign. That is, there are no dramatic costs for missteps. However, while there are no dramatic consequences for bad actions, faults must be corrected fairly quickly, since otherwise failure costs continue to accrue. Second, especially for D-IPI, the cost/failure curves holding number-of-steps fixed are not generally monotonic. This implies that practical use of this algorithm might require careful "tuning" of the number of steps. We are planning to explore strategies in which the number of steps is context dependent. Finally, both D-IPI and PK-reduced substantially outperformed both the random strategy and exact computation over the entire range of our experiments.

D-IPI and PK-reduced were the clear winners in this set of experiments. PK-reduced, in particular, provided very smooth and consistent performance over the entire experimental range, and required less tuning than D-IPI. On the other hand, K-reduced performed reasonably well, but its performance was not terribly predictable, and it performed very poorly in the early stages of computation.

### Scaling

These results are interesting, but it is difficult to generalize from a single data point. Three questions are of interest:

    1. How do the results scale with increasing problem size?

2. How do the results scale with increasing look-ahead depth (number of decision stages)?

3. How are the results affected by parameters of the particular problem (failure probabilities, replacement costs, inspection costs, etc.)?

We are beginning to explore these issues, and have some preliminary data on the first two questions.

### Scaling with Problem Size

In earlier work with D-IPI we tested scaling with problem size by building a series of test problem instances of increasing size [D'Ambrosio, 92]. The problem instance used in this paper, a 4-gate "half-adder" circuit, was drawn from the middle of that series, in which problem instances ranged from one to sixteen gates. In that study we found that both posterior estimation and decision evaluation times, for small numbers of steps, grew only slightly faster than linearly with number of gates.

### Scaling with Look-ahead Depth

A recent study by one of us (D'Ambrosio) at Prevision looked at evaluation complexity of multistage decision problems as a function of both the number of steps of computation and the number of decision stages. The problem studied was target identification. In the variant we studied, a single platform (aircraft) would be detected at a random distance, moving directly towards the agent. The agent had available a number of noisy sensors, and its goal was to "declare" the identification of the target as soon as possible. Sensors had varying costs per use, and improved in reliability as the target approached. The utility of declaring the correct identification declined as the target approaches, and there was a substantial disutility for incorrect declaration.

The table below shows the inference time required per decision by D-IPI, in cpu seconds for Common lisp on a Macintosh Quadra 610.

| Depth | Steps: 1 | 3 | 6 | 10 | 16 |
|---|---|---|---|---|---|
| 1 | 2.1 | 3.7 | 4 | 7 | 16 |
| 2 | 3 | 3.5 | 4.5 | 15.5 | 20.6 |
| 4 | 3.5 | 3.5 | 6.2 | 15.5 | 25 |
| 5 | 6.5 | 6 | 9.2 | 19.5 | 45 |
| 6 | 6.7 | 6.7 | 14 | 31 | 64 |
| 7 | 10.5 | 14 | 23 | 24.5 | 87.5 |
| 8 | 10.5 | 9.7 | 15.7 | 30 | 116 |
| 9 | 11 | 10.5 | 19.2 | 37 | 108 |
| 10 | 19 | 25 | 55 | 35 | 120 |

Table 7: Inference Time Per Decision

These results are surprising. We did not expect IPI to be able to search to depth 10 without incorporation of significant domain heuristics. We did include one simple heuristic in the search: we ruled out consideration of declaration acts for any target id other than the actual id hypothesized in the current scenario[5]. This heuristic did not require modification of the algorithm since it can be expressed as a local expression [D'Ambrosio, 91] on the decision node domain.

In contrast, our experiments revealed a problem in applying PK-reduced to the single-target ID problem. As shown in the table below, PK-reduced is intractable for depth greater than 3. We believe these results are due to the fact that PK-reduced restricts domains statically, rather than dynamically. As a result, it must consider the full cross product of the restricted domains, a phenomenon to which D-IPI is not subject.

| Depth | Steps: | 4 | 16 | 64 |
|---|---|---|---|---|
| 2 | 2 | 3.6 | 4.5 | 6 |
| 3 | 9.8 | 25 | 65 | 67 |
| 4 | 35 | 61 | | |
| 5 | 113 | | | |

Table 8: Evaluation Complexity of PK-reduced

### Sensitivity to other parameters of the decision model

Both D-IPI and PK-reduced depend on skewness in the given probability and utility distributions. Without this, neither can be expected to perform well. However, not all of the distributions in our test problem satisfy the "skewness" criterion in [D'Ambrosio, 93]. Each of the gates has an *unknown* mode in which its output distribution is uniform.

Further, it is not obvious why they should perform well in a decision context, even when all distributions are skewed. For example, even when a few high-utility scenarios contribute the bulk of expected utility to each decision alternative, it is not clear that the remaining scenarios might not contribute enough mass to change the decision. They apparently do, at least some of the time: if this were not the case, performance in the OLMA problem would never improve with increasing computation. However, our experimental results indicate that while performance starts out quite good, it does in fact improve with increased computation for both D-IPI and PK-reduced. Further study is needed to better understand the conditions that enable this.

---

[5] Remember that IPI is a search-based algorithm that proceeds by instantiating variables in the network - we call each such instantiation a *scenario*.

## Related work

[Draper & Hanks, 94] investigate *localized partial evaluation* of Bayesian belief networks to perform anytime inference. [Poole, 93] has done work on the use of conflicts for reducing necessary computation, and [Wellman & Liu, 94] have applied state space abstraction to address resource-bounded computation. While the ideas are promising, further empirical validation is necessary to demonstrate that these techniques are scalable to and competitive on large, general problems.

The OLMA may be viewed as a POMDP. Many researchers in machine learning seek optimal or near-optimal policies for POMDPs using variations on value iteration and Q-Learning [Parr & Russell, 95; Jaakkola, Jordan and Singh; Littman, Cassandra, & Kaelbling, 95]. Finding optimal policies for models requiring even tens of states currently stretches the limits of feasible computation [Parr & Russell, 95]. Still, these papers demonstrate a marked improvement in the ability to calculate optimal policies. Closer to home, we have begun to investigate POMDP methods for the OLMA domain [D'Ambrosio, NIPS96-submitted].

Most closely related to our work are examinations of resource-bounded algorithms for belief networks. [Horvitz et al., 89] employs *bounded conditioning*, a technique we believe may perform well in the OLMA and which we hope to include in some future investigations. We likewise will seek competitive forms of stochastic simulation [Fung & Chang, 89], and continue our explorations with the kappa calculus [Goldszmidt, 95].

## Conclusions

We are interested in developing and characterizing decision algorithms with robust real-time performance. We presented the On-Line Maintenance domain, a domain we think is uniquely suited for effective evaluation of real-time decision methods. We then presented preliminary results indicate that two algorithms, D-IPI and PK-reduced, exhibit the tradeoff between computation time and decision quality necessary for good performance in this test domain.

## Acknowledgments

This work done with the support of NSF grants IRI-950330 and NSF CDA-921672.